# An innovative solution for breast cancer textual big data analysis

N. Thiebaut[1], A. Simoulin[1], K. Neuberger[1], I. Ibnouhsein[1], N. Bousquet[1,2] *, N.Reix[3,4], S. Molière[5] and C. Mathelin[6,7,8]

[1] Quantmetry, 128 rue du Faubourg Saint-Honoré, 75008 Paris, France.
[2] Université Pierre-et-Mare Curie, 4 place Jussieu, 75005 Paris, France.
[3] ICube UMR 7537, Université de Strasbourg / CNRS, Fédération de Médecine Translationnelle de Strasbourg, 67200 Strasbourg, France.
[4] Laboratoire de Biochimie et Biologie Moléculaire, Hôpitaux Universitaires de Strasbourg, 1 place de l'Hôpital, 67091 Strasbourg, France.
[5] Department of Imaging, Hôpitaux Universitaires de Strasbourg, 1 avenue Molière, 67098 Strasbourg, France.
[6] Unité de Sénologie, Hôpitaux Universitaires de Strasbourg, 1 avenue Molière, 67098 Strasbourg, France.
[7] Department of Functional Genomics and Cancer, Institut de Génétique et de Biologie Moléculaire et Cellulaire, CNRS UMR 7104, INSERM U964, Université de Strasbourg, Illkirch, France.
[8] Centre Hospitalier de Sarrebourg, rue des Roses, 57400 Sarrebourg, France.

**Abstract**
**Motivation:** Traditional methods in epidemiology are based on the development of cohorts of individuals which are monitored during a certain period of time. The digitalization of stored information in hospitals now allows for the exploitation of medical data initially gathered for other purposes than epidemiology. More specifically, hospitals continuously gather huge amounts of textual data in electronic health records (EHRs). Storing health data in text format is convenient, but without processing, manual search and analysis operations on such data become tedious. Automating the extraction of information from vast amounts of EHRs would provide new data sources for epidemiology and other research fields. In recent years, different studies reported the use of natural language processing (NLP) tools to structure the information contained in electronic health records and perform statistical analysis on this structured information. The main difficulties with the existing approaches is the requirement of synonyms or ontology dictionaries, that are mostly available in English only and do not include local or custom notations. Therefore, we experimented the creation of a team composed of oncologists as domain experts and data scientists to develop a custom NLP-based system to process and structure 14,029 textual clinical reports from 2000 to 2017 of patients suffering from breast cancer.
**Results:** The tool relies on the combination of standard text mining techniques and an advanced synonym detection method. It allows for a global analysis by retrieval of indicators such as medical history, tumor characteristics, therapeutic responses, recurrences and prognosis. The versatility of the method allows to obtain new indicators with ease, thus opening up the path for retrospective studies with a substantial reduction of the amount of manual work. With no need for biomedical annotators or pre-defined ontologies, our structuring method reached an extraction accuracy that ranges from 70 to 96.8 % for several concepts of interest, according to a comparison with a manually structured file. Unlike existing method, the text structuring technique presented in this paper is language agnostic, and does not require to complete an existing corpus with local or new notations.

**Contact:** nbousquet@quantmetry.com

## 1 Introduction

In the last two decades, the available storage capacities increased exponentially, thus leading to bigger volumes of stored medical data. This medical data is very heterogeneous in nature: images, videos, audio records, genomics, structured data or free text are just a few examples of data types daily manipulated in medical centers. Generally, these data are exploited at an individual level: for instance a Magnetic Resonance Imaging (MRI) or a free text health record will be analyzed to establish the diagnosis or follow the disease evolution for a specific patient. Among all the different data types, only structured data can be readily exploited for statistical purposes, which include, for instance, the detection and explanation of patterns.

However, in medical centers, an important part of useful medical information is stored in unstructured free text format, often as electronic health records (EHRs). For instance, many hospitals report the follow-up of patients suffering from breast or other cancers in multidisciplinary meeting notes. The unstructured nature of this data makes manual search and analysis at a meta-individual level tedious and time consuming; hence its use is mostly restricted to individual monitoring purposes. However, these data represent a rich source of information for studies on populations that requires the application of automatic pattern extraction techniques on EHRs, based on deep learning[1,2] and more specifically on a class of "natural language processing" (NLP) algorithms[3,4]. While many successful experimentations in psychiatry[5], oncology[6,7], breast cancer study[8—12] and admission pattern reconstruction[13] have been reported, most of these studies use specialized ready-to-use tools to search for a specific information, for instance cancer recurrence. These tools are usually field-specific, and impose a standard vocabulary that is not always used in EHRs. The construction of these annotation tools requires a large amount of manual work by domain experts. In addition to this vocabulary normalization problem, medical health records often include abbreviations, misspellings and a vocabulary that is local and evolves with time. While these tools are very helpful for exploiting EHRs, clinicians still lack a versatile tool that allows to target any type of information in these documents to allow for statistical studies or comprehensive searches.

We report in this article the development of an innovative solution based on NLP techniques that allows for a multi-targeted analysis of free text medical data. To build the search engine, we used 14,029 anonymized multidisciplinary meeting notes corresponding to 9,599 breast cancer patients collected between 2000 and 2017 by the senology department of the Hautepierre Hospital in Strasbourg, France. The notes were written in French, but since we have not used any external corpus our methodology can be used to harness other languages or chronic diseases equally easily.

The technical core of the tool is based on recent automatic synonyms detection techniques, an algorithm for typographical errors corrections, and repeated interactions within an interdisciplinary team composed of oncology domain experts and data scientists. Contrary to other structuration methods, the synonyms base is built based on the used corpora, which makes all the process easily replicable and adaptable to any local language specifications. We could retrieve several indicators such as tumor sizes, tumor types, hormonal responses or presence of biomarkers to produce various statistical studies on populations. Moreover, medical practices such as the RMI precision or hospitalization duration can be investigated. The methodology reported in this paper may help better understand cancer and improve health care in this field.

## 2   Methods

Text files structuring is a challenging task, mostly because many different formulations may refer to the same concept in a given context. Indeed, a text-mining study has reported 124 alternative formulations for "ductal invasive carcinoma" in 76,000 breast pathology reports[7]. To handle this complexity, we have developed an approach to reduce the semantic complexity of documents, thereby making it workable for search and information retrieval algorithms. To do so we have created a synonym dictionary specific to our corpus using the recent Word2Vec algorithm[14] ; this dictionary captures the various abbreviations and formulations that are corpus-specific. This synonym detection step is crucial to the performance of our information retrieval process, as many formulations are either local or specific to the annotator. Hence, with the sole use of a standard medical dictionary we could not retrieve information such as the cancer type from text files. The text transformation process and the synonym detection algorithm are described below (Fig. 1).

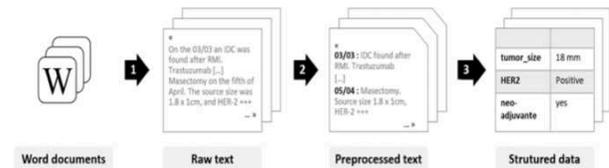

**Figure 1: Text structuring process**. Our input data is a collection of EHRs stored as Word Documents. The first structuring step consists in extracting the text, keeping sections split but getting rid of uninformative text formatting. The second structuring step performs text normalization (accents removal, words stemmification, etc.) and chronological information extraction, in order to know the chronological order of the various operations and events. The third and last step of our structuring process outputs structured data, with relevant fields extracted from textual data.

Our input data corresponds to multidisciplinary meeting notes in text format (Word files). We proceed in three steps to extract information patterns from free text:

1. An integration step, where raw data is transformed into the so-called JSON format with sections split and general information such as age and date retrieved.
2. A preprocessing step, where first text mining treatments such as accents removal are applied, and the medical history is reconstructed using dates retrieval, chronology reconstruction, and location identification (left, right or both breasts in our case).
3. A structuration step that involves a rule engine to partially automate local human expertise and a synonym detection step based on the "Word2Vec" algorithm[14], in order to obtain a comprehensive set of formulations for each concept under investigation

Once information is extracted, it is structured in a table format, ready to be used for statistical studies.

### 2.1   Preprocessing

Text structuring requires several preprocessing steps in order to reduce the unnecessary diversity of words occurrences and phrases constructions.

- **Accents and other language-specific symbols:** like for most NLP tasks, we could remove language specifics without creating ambiguities.
- **Plural and conjugations:** standard normalization techniques consist in the truncation of the plural signs, and the replacement of words with their lemma. The latter methods happen to destroy too much information in our data, thus preventing a reliable analysis.

Therefore, we have not performed any stemming or lemmatization processing.
- **Chronology reconstruction:** finding dates and chronology indications helped us to find, for instance, neo-adjuvant treatments. This step is achieved with simple dates parsing.
- **Typographical errors corrections**: a typographical error often appears as a word with rare occurrences that is similar to another word that is more frequent. For all rare words, for which the number of occurrences in the corpus is lower than a configurable threshold, we check if there is a similar word that is more frequent and replace the former with the latter if so. Technical details about the choice of metric and threshold are provided in Supplementary Online Material.

For example, the text "dettection of multiple tumors on July, 06th" would become "detection of multiple tumors <date>" once all the above mentioned preprocessing steps are performed, the date being extracted and stored in a normalized form.

## 2.2 Text files structuring

### 2.2.1 Text standardization and synonyms detection

As mentioned in the introduction, various synonyms or abbreviations referring to the same concept have to be gathered under a single term. For instance, with this structuring strategy "infiltrating ductal carcinoma" and "invasive ductal carcinoma" are both replaced by the acronym "IDC". Notice that an additional complexity arises from the fact that synonyms may be composed with a different number of words. We resolve this issue by grouping all pairs and triplets of words that frequently appear together (see the Supplementary Online Material for details on the grouping strategy).

Synonyms detection algorithms search for words sharing similar contexts (surrounding phrases). Those algorithms cannot perfectly discriminate antonyms and synonyms because both might appear within the same context. Thus, a manual verification by members of the medical staff is needed (see Fig. 2).

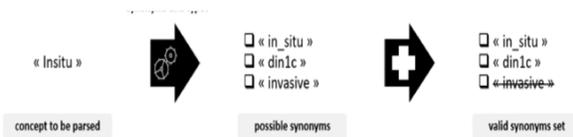

**Figure 2: Concepts parsing process**. The medical staff defines the name of the concept to be parsed, and possibly a few of its synonyms or alternative formulations. The synonyms detection program then finds all similar words in the corpus. The latter potential synonyms are validated by the medical staff to avoid antonyms/synonyms mixing. Frequent typographical errors are also corrected in the process, being detected by the algorithm as "synonyms".

The named entity recognition task in medicine usually relies either on the use of existing corpora like the ones provided by the National Library of Medicine's Unified Medical Language System (UMLS)[8,15], on machine learning systems trained using manually annotated texts[16], or on a combination of the latter methods[17].

While the first method may be very efficient, corpora based methods are not able to deal with local abbreviations or notations. Also, most resources are only available in English and thus are not suited for other languages. Although medical corpora exist in French[18], the typical precisions reached for NLP tasks with those corpora are significantly lower than in English (83% provided by Névéol and her coauthors[18]), due to the smaller number of contributions.

The second named entity recognition method consists in manually annotating part of the corpus being studied, so that machine learning algorithms can be trained to automatically recognize concepts and ontologies on the whole corpus. This method is highly adaptive and can accommodate the notations peculiarities specific to an institution or a writer. Nevertheless, a large volume of annotated documents is necessary to train an accurate named entity recognition program based on machine learning. The manual annotation, albeit fruitful, is both prone to error and time-consuming.

In this work we propose a novel named entity recognition method for medical corpora that relies on a recent algorithm called Word2Vec[14]. With this method words are mapped to vectors, the mapping being learnt in an unsupervised way on the studied corpus, such that words with similar meanings are mapped to close vectors. For each concept to be parsed the algorithm proposes a set of synonyms that correspond to the closest vectors in the Word2Vec vector space. Nevertheless, the algorithm sometimes suggests wrong synonyms that are most often antonyms or words occurring in similar sentences. For instance, type I diabetes may be proposed by the algorithm as a synonym for type II diabetes because they occur in similar sentences, while they should evidently be mingled. For this reason, we have worked with the medical staff to validate the synonyms proposed by the program in order to retrieve abbreviations, acronyms, synonyms or formulations that might have been forgotten when the synonyms dictionary was first built. More detailed explanations about the Word2Vec algorithm are provided in the Supplementary Online Material.

Our medical records structuring process is iterative: for each concept under scrutiny, a first set of synonyms is provided by the medical staff. We use this first set as an input for the synonym detection algorithm, that outputs additional possible synonyms to be checked by the medical staff. This process is continued until the synonym detection program is not able to propose new synonyms that are considered valid by the medical staff (see Fig. 3).

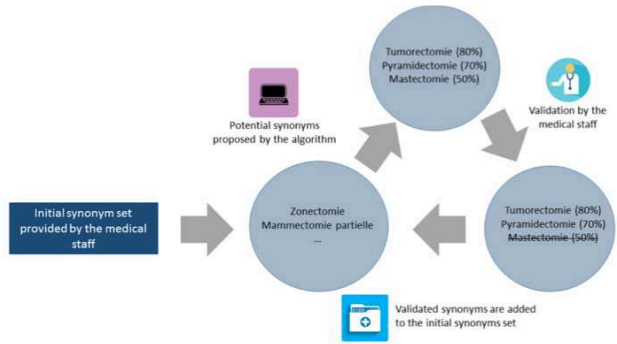

**Figure 3: Iterative process for creating a dictionary**. For a given concept ("cancer" in this example), a list of synonyms is provided by the medical staff. The synonym detection algorithm searches for other synonyms. The list proposed by the algorithm is then validated by the medical staff. Finally, the initial set of synonyms is augmented with the synonyms proposed by the algorithm and validated by the medical staff. The process is repeated until no additional synonym is found.

#### 2.2.2 Text files structuring

Once all the text standardization and synonyms detection steps are completed, extracting structured information from textual data is straightforward. But in some cases, we want to detect more information than the mere presence or absence of a concept. For instance, if the tumor concept is detected, we generally want to retrieve the tumor size. For the latter example and for others such as hormonal receptors responses, we have implemented a few elementary parsing rules to retrieve the relevant information (often the selection of a block of words followed by a search for specific terms).

#### 2.2.3 Structuring improvement with multiple textual data sources

In order to improve the rate of records for which we extract tumors and patients' characteristics successfully, we have applied the same structuring process to the 7,360 hospitalization letters and 3,405 discharge letters at our disposal for the same period and patients. Those letters, written by the medical staff and addressed to the patient's regular doctor, are stored by the hospital. They summarize diagnoses, operations and various other information about the patient. Missing or incorrectly written information in the medical records of a patient can be found in the letters, thus effectively improving our extraction rate.
Similarly, all textual data sources can be combined to improve extraction or increase the number of characteristics extracted.

### 3 Results

#### 3.1 Performance evaluation: Comparison of our text structuration with manually structured corpora

In order to estimate the precision of our structuring methodology, we compared the extracted values with two different logbooks maintained by the Hautepierre hospital. Several characteristics of 322 of our patients were present in the first logbook in a structured form, such that for each case a direct comparison of our program output and the logbook register was possible.

The results of this comparison are presented in Table 1. Agreement ratio varies between 89.5 % and 98.3 % depending on the indicator. These agreement scores should be compared with typical inter-annotators agreement which is usually close to 95 %[16,18].

**Table 1**. Comparison of 5 indicators extracted with our method and the structured data available for a randomly sampled subgroup of 322 individuals from our cohort. The agreement A between the structured data and the output of our program is shown in percent. The precision P, sensibility S and $F_1$ score (which is the harmonic mean of precision and recall) are shown for categorical variables. The extraction rate ER is the ratio of medical records for which the indicator was found.

| Indicator | A (%) | P (%) | S (%) | $F_1$ score (%) | ER (%) |
| --- | --- | --- | --- | --- | --- |
| SBR stage | 89,5 | - | - | - | 99,0 |
| Estrogen receptor | 98,3 | 98,5 | 99,6 | 99,0 | 94,5 |
| Progesterone receptor | 92,5 | 91,5 | 99,1 | 95,1 | 94,5 |
| HER2 | 95,9 | 80,0 | 60,0 | 68,6 | 86,8 |
| Number of metastatic lymph nodes | 90,1 | - | - | - | 84,8 |

EHRs that produced disagreement between our information extraction method and the structured data file were examined manually for the number of metastatic nodes. In 13 of the 17 discordant values, manual review validated the values found with our method. The four other cases correspond to peculiar situations that the natural language processing algorithm was not able to handle correctly.

The second logbook we used to assess indicators retrieval quality contains structured information on the hormonal receptors and SBR stage (like the first logbook), alongside information on the rate of cell growth (percent of tumor cells positive for the Ki-67 protein) and cancer type (ductal, lobular, etc.).

As for the first logbook, values extracted with our method were compared with manually extracted ones for the 65 patients present in the second logbook. For this group, the agreement varies between 90.0 and 96.8 % depending on indicators.

In addition to the correctness of indicators for a given subpopulation, we have also evaluated the capability of our method to retrieve the relevant medical records for given concepts. The second logbook contained only patients who had worn a Levonorgestrel Intra uterine device (IUD). We have checked that we could extract the medical records of the same patients with our built search engine, searching for the word Levonorgestrel IUD and its synonyms in the preprocessed and typographical errors-free corpus. Eventually, we missed one patient and found one additional relevant patient that was absent from the logbook.

Hence, as illustrated by this example, our method is as efficient as manual extraction for the identification of subpopulations with common characteristics.

**Table 2**. Comparison of the indicators extracted with our method and the structured data available for a randomly sampled subgroup of 65 individuals from our cohort. A is the agreement between the structured data and the output of our program (shown in %), κ is Cohen's Kappa, P is the precision (%), S is the sensibility (%) and ER is the extraction rate (%). Cohen's Kappa κ is shown for only the cancer subtype indicator because it is a relevant agreement measurement for multi-class variables, and it does not depend on the proportion of the different cancer subtypes.

| Indicator | A (%) | κ (%) | P (%) | S (%) | $F_1$ score (%) | ER (%) |
|---|---|---|---|---|---|---|
| SBR III stage | 90,6 | - | 96,2 | 83,3 | 89,3 | 100,0 |
| Estrogen receptor | 93,4 | - | 97,7 | 93,3 | 95,5 | 93,8 |
| Progesterone receptor | 90,0 | - | 88,9 | 97,6 | 93,0 | 92,3 |
| Ki-67 | 91,4 | - | - | - | - | 73,8 |
| Cancer subtype (ductal, lobular, …) | 96,8 | 91,5 | - | - | - | 96,9 |

### 3.2 Synonyms detection algorithm contribution and effectiveness

Among all the indicators we have extracted from our text corpus, some indicators such as the hormonal receptors and tumor sizes were obtained by the sole application of text extraction rules (regular expressions), and the synonyms dictionary we used was entirely provided by the medical staff. For some other indicators, the dictionary had to be supplemented with synonyms detected by the Word2Vec based algorithm. For instance, for the SBR stage we could find that the word "grade" could also be written as simply "gr" or "g", using the synonyms detection algorithm. Thus, we were able to find the SBR stage in more medical records with those additional synonyms.

The results presented in Table 3 indicate the contribution of the synonyms detection algorithm for the parsing of some indicators for which it led to an improvement. This algorithm allowed us to establish an exhaustive synonyms list for a concept by finding synonyms, notations or formulations that were used scarcely or temporarily in our text corpus.

**Table 3**. Contribution of the automated synonyms detection program (based on the Word2Vec algorithm) for a few concepts for which it leads to an improvement. We show the parsing ratio (proportion of the medical records where the information was retrieved) for three indicators with and without the synonyms suggested by the synonyms detection program.

| Indicator | Parsing ratio (%) without synonyms detection |
|---|---|
| SBR stage | 64.80 |
| Cancer type | 96.55 |
| Cancer subtype (ductal, lobular, …) | 87.17 |

### 3.3 Applications

Our structuring method empowers the medical staff with a search engine that can find phrases in the corpus, taking into account synonyms and alternative formulations to retrieve medical records with specific characteristics. This allows to find the medical records of all patients with a specific feature, and compare their tumoral characteristics with those of the rest of the cohort. We have already applied this process to compare the tumoral characteristics of:

- The patients who had a Levonorgestrel IUD with the rest of our cohort;
- Elders and younger women;
- Type 2 diabetics and non-diabetics ;
- Small tumors.

As an example of possible statistical studies, medical records that mention diabetes mellitus were found and isolated with our search engine. The structuring program then allowed for a direct comparison of the characteristics of breast cancer in case or in absence of type 2 diabete. We have found 7,071 patients older than 50 years old with invasive tumors and no diabete, and 473 similar patients with type 2 diabete. We have found statistically significant differences in the SBR grade (p=0.2%), estrogen receptors response (p=1.6%) and tumor sizes (p=0.1%).

**Table 4**. Comparison of various indicators for patients above 50 years old suffering from diabetes mellitus, and patients within the same age range but without type II diabete.

| Indicator | Non-diabetic | Type 2 diabetics |
|---|---|---|
| Average tumor size | 22,4 mm | 26,5 mm |
| Positive estrogen receptors ratio | 85,5 % | 87,5 % |
| Positive progesterone receptors ratio | 73,3 % | 76,7 % |
| Positive HER2 ratio | 12,4 % | 9,8 % |
| SBR I grade ratio | 29,2 % | 22,1 % |
| SBR II grade ratio | 45,0 % | 46,3 % |
| SBR III grade ratio | 25,8 % | 31,5 % |
| Metastatic axillary lymph nodes | 37.77% | 40.61% |

## 4 Discussion

We presented a new methodology for structuring data and searching cases in a vast corpus of unstructured medical textual data. The innovation stands on the iterative entanglement of automated tasks and medical feedback. The latter reduces errors an inserts medical expertness within the procedure. Both the search engine and the structuration process are able to handle the typographical errors, negations, and other sources of complexity in the French language thanks to a combination of various natural language processing techniques. Our program used a dictionary of synonyms provided by the medical staff, which was eventually augmented with an innovative synonyms detection technique that relies on the recent Word2Vec algorithm. This methodology allowed us to obtain good information retrieval results by comparison with structured logbooks available for a subgroup of our cohort.

Our method does not require any external corpora and can then be used for any language. Also, it does not require manual annotation of text for the purpose of pattern recognition learning like supervised learning methods do. This setup necessitates frequent interactions between the medical staff and the data scientists in charge of the data analysis, but the time required was very limited compared to a manual annotation of text files.

The semantic complexity of medicine was studied for the case of breast cancer by Xu and his coauthors[4]. In this work, in more than 76,000 breast pathology reports, 124 alternative formulations were found for "ductal invasive carcinoma", 95 for "lobular invasive carcinoma", and more than 4,000 formulations had to be interpreted as an absence of ductal invasive carcinoma. This richness in formulation equivalences is pervasive in medicine, and supports the necessity for the use of advanced natural language processing techniques. More generally, our work demonstrates the possibility to study diseases and medical practices using unstructured data. Nowadays, this type of data is massively present in hospitals databases, but is still mostly unused.

Moreover, the structuring process produces tabular data files for which translation is straightforward, and thus allows for comparisons of pathologies in different countries. While most studies relying on textual data are led in English-speaking countries, our work allows to reproduce the same studies in different countries, and could highlight differences between populations.

One of the main advantages of our structuring process is the possibility to search for potential risk factors, sub-populations specificities or cancer types peculiarities a posteriori with almost no additional work. This work could be applied to any pathology with the same process, and just requires the regular textual data that are stored in all hospitals. The application of this methodology to specific studies will be the subject of future articles.

Beyond the strict structuration of textual data, the application and extension of this methodology could lead to improve a prognostic by merging pieces of information of various nature and supports. In oncology, a prognostic of tumoral aggressiveness based on reports of multidisciplinary team meetings (MDTM), discharge letters and textual information about sociological factors[19], among others, could be significantly reinforced using phenotypic data resulting from clinicopathologic examinations, metabolic data[20], microbiological data[21] including histological type or the presence of therapeutic targets (HER2, hormone receptors) and imaging data. Improving the sensitivity of mammography-echography imaging[22,23], the use of 3D magnetic resonance imaging requires to handle massive datasets, which reflects the scalability challenges to be encountered in extension studies. Future work will be dedicated to building such a methodology and, again, applying it to answer specific medical questions or problems.

A first step towards such a methodology was already carried out by several of the authors on a subset population of 2139 female patients, involving follow-up and discharge letters, reports of MDTM and reports issued by breast-screening institutions, confirmed or completed by billing codes[24]. Statistical studies resulted in highlighting tumor and therapeutic characteristics. Furthermore, they highlighted that the breast-screening French procedure, based on regular mammograms possibly completed by ultrasounds, truly improves the proportion of cancers detected at an early stage (which avoids heavy treatments and leads to a reduction of mortality). Such a result provides a quantitative answer to a major public health problem, by providing arguments against the growing skepticism towards the efficiency of the procedure regularity[25]. This dramatically illustrates the relevance and usefulness of the methodology presented in this article.

## Acknowledgements

The authors thank Romain de San Nicolas (Quantmetry) for his support and management during the research work which allowed to produce this article.

## Funding

This research was funded through a local call for projects. ClinicalTrials.gov Identifier: NCT02810093.

*Conflict of Interest:* none declared.

# Supplementary material for the article "An innovative solution for breast cancer textual big data analysis"

N. Thiebaut[1], A. Simoulin[1], K. Neuberger[1], I. Ibnouhsein[1], N. Bousquet[1,2] *,
N.Reix[3,4], S. Molière[5] and C. Mathelin[6,7,8]

[1] Quantmetry, 128 rue du Faubourg Saint-Honoré, 75008 Paris, France.
[2] Université Pierre-et-Mare Curie, 4 place Jussieu, 75005 Paris, France.
[3] ICube UMR 7537, Université de Strasbourg / CNRS, Fédération de Médecine Translationnelle de Strasbourg, 67200 Strasbourg, France.
[4] Laboratoire de Biochimie et Biologie Moléculaire, Hôpitaux Universitaires de Strasbourg, 1 place de l'Hôpital, 67091 Strasbourg, France.
[5] Department of Imaging, Hôpitaux Universitaires de Strasbourg, 1 avenue Molière, 67098 Strasbourg, France.
[6] Unité de Sénologie, Hôpitaux Universitaires de Strasbourg, 1 avenue Molière, 67098 Strasbourg, France.
[7] Department of Functional Genomics and Cancer, Institut de Génétique et de Biologie Moléculaire et Cellulaire, CNRS UMR 7104, INSERM U964, Université de Strasbourg, Illkirch, France.
[8] Centre Hospitalier de Sarrebourg, rue des Roses, 57400 Sarrebourg, France.

**Abstract**
**This material provides additional information to the articleon some operations of complexity reduction conducted during textual analysis.**

## 1 Vectorization of words and reduction of text complexity using Word2Vec

It is part of our methodology to reduce the vocabulary variety by gathering synonyms and abbreviations under a single term. This operation is not straightforward as the words themselves do not carry information about their synonyms. Indeed, two words with a close sense might not have a close spelling. For example, the "amplification" test and the "SISH" test refer to the same concept. However, the spelling, *i.e.* the representations of those words are completely different. Therefore it is useful to have an embedding representation of words, such that two words with a close sense have a close representation. Algorithms that learn meaningful representations from a dataset of texts belong to the so-called *word embedding* methods[1,2]. While usual words can be represented as mathematical elements (vectors) defined in a high dimensional space, but submitted to *sparsity* (namely, with many null coordinates), word embedding provides a representation $v$ of a word $w$ as a dense vector in a lower dimensional (output) space $E$ of dimension $d$, where $d$ is typically in the order of magnitude of a few hundred.

This embedding space $E$ offers multiple useful properties. All kind of semantic relationships are captured in the representation and can be retrieved with linear transformations as shown in Fig. 1.

New methods recently emerged to build a transition function from the original ("one-hot-encoded") representation of words to the embedding space $E$. In particular, the Word2Vec algorithm proposes to produce this transformation using a neural network architecture[1,2] (Fig. 2). The idea behind this algorithm is to ensure that words that usually appear in the same context are close in the output representation space, where the context is defined as the ensemble of surrounding words, typically the five preceding and following words. The Word2Vec algorithm has two variants: the CBOW and Skip-Gram implementations, whose objectives are respectively to predict a central word given his context, or conversely to predict the context given a central word.

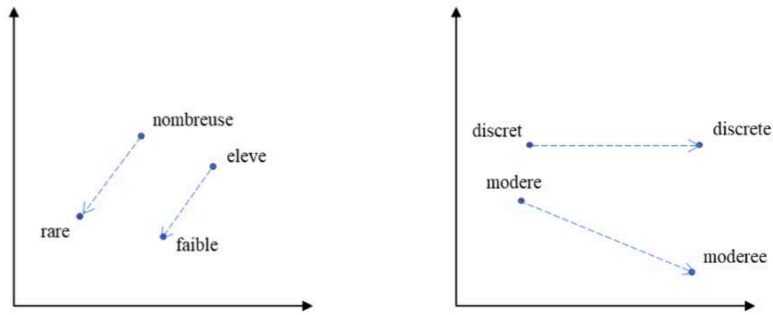

**Fig. 1.** The Word2Vec algorithm was applied to our corpus. The resulting representation of the 1000 most frequent words in the corpus was projected into a two-dimensional space using the t-distributed stochastic neighbor embedding algorithm[3] (t-SNE). In the word embedding space, the same distance can be observed between the pairs ("nombreuse","rare") ("numerous" and "rare" in English) and ("eleve","faible") ("high" and "low"). This illustrates the fact that the representation learnt by the word2vec algorithm captures semantic relationships such as antonyms and feminine forms.

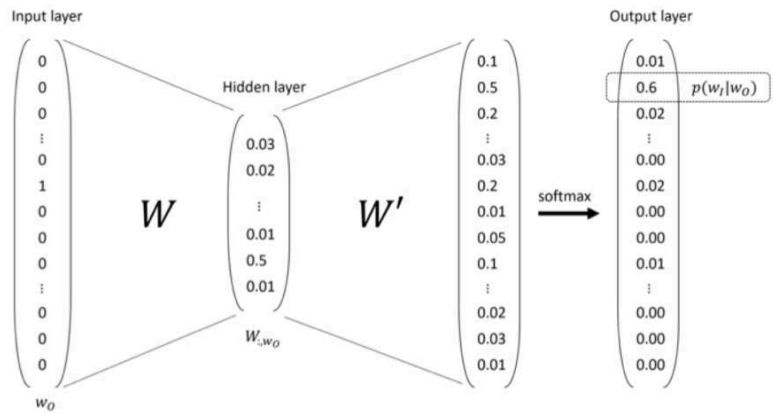

**Fig. 2.** Word2Vec neural network representation. In the input layer a word w is represented by a one-hot-encoded vector (0, 0, …, 0, 1, 0, …, 0) that only contains zeros except at the (arbitrary) index of the input word. From the one-hot-encoded input vector x, the Word2Vec vector representation h is given by the product of a matrix W and x. Finally, the Output Layer contains the probability distribution of the predicted words, and is obtained by applying a softmax function to the product of h with a second matrix W'. The coefficients of W and W' are both learnt by gradient descent.

In the Skip-Gram implementation, the objective is to predict the context, *i.e.* the surrounding words of a given word. Every word $w$ in the corpus is projected in the embedding space by using a transformation matrix $W$. The word representation is then the vector $v_w = W \cdot w$. This representation is then projected back in the original word space by using a matrix $W'$. The results of this second projection are transformed into a probability vector using a softmax function. The probability $p(w_I|w_O)$ for a given words $w_I$ to appear in $w_O$ context is finally given by:

$$p(w_I|w_O) = \frac{\exp(v'_I{}^T v_O)}{\sum_{w=1}^{|V|} \exp(v'_w{}^T v_O)}$$

where $|V|$ is the vocabulary size of our corpus. The parameters of the two projections matrices $W$ and $W'$ are learnt with gradient descent by minimizing the average log-probability $J(W, W')$, defined as:

$$J(\theta) = \frac{1}{T} \sum_{t=1}^{T} \sum_{-c \leq j \leq c, j \neq c} \log p(w_{t+j}|w_t)$$

Below is presented a two-dimensional representation of the 1000 most frequent words in our corpus. The visualization illustrates the formation of cluster gathering words with close meaning.

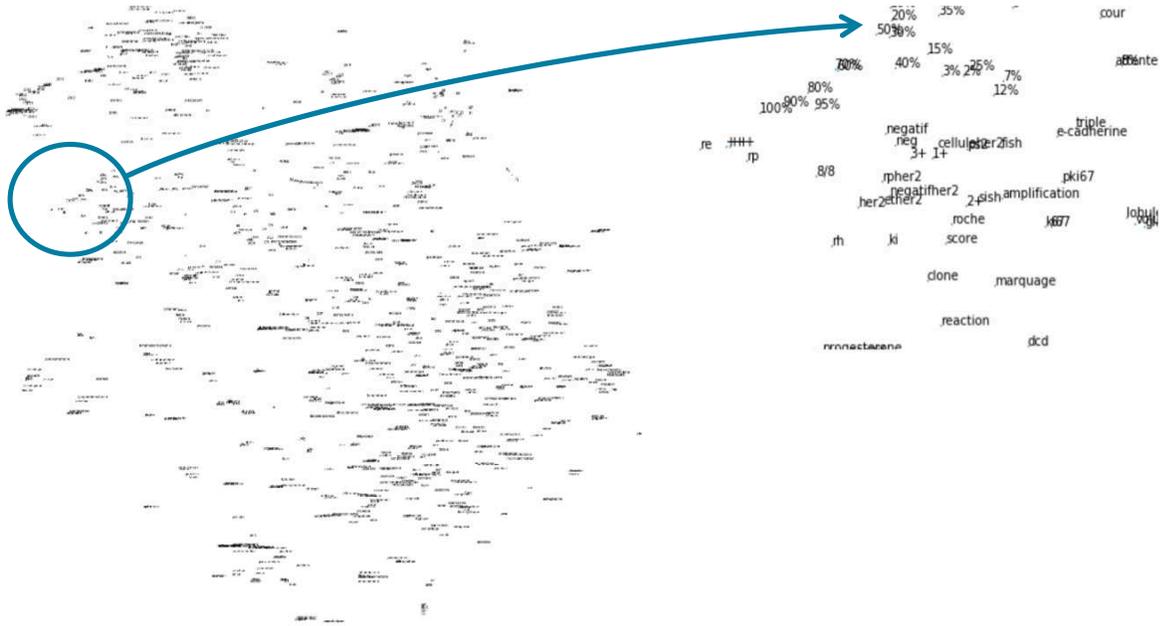

**Fig. 3.** (Left) The Word2Vec algorithm was applied to our corpus. The resulting representation of the 1000 most frequent words in the corpus was projected into a two-dimensional space using the t-distributed stochastic neighbor embedding algorithm (t-SNE). (Right) Portion of the cloud presenting words associated with the vocabulary referring to the histological type of tumor: HER2, KI67, RE, RP.

## 2   Phrases detection and pointwise mutual information

Phrases detection is a common and important preprocessing step in natural language processing. Indeed, for some applications it is critical to treat successive words as a single entry. For instance, in our case we had to distinguish the phrase "partial mastectomy" from the word "mastectomy" in the breast cancer multidisciplinary meeting notes at our disposal, as those two formulations refer to two different operations. In order to find patients who underwent total mastectomy, a simple search of the word "mastectomy" would also output patients who underwent partial mastectomy. Hence phrase detection is necessary to distinguish between isolated occurrences of a word and meaningful phrases.

We have borrowed the simple phrase detection method from Mikolov and his coauthors[2] where pairs of successive words $w_i$ and $w_j$ are added to the vocabulary as bigrams if their relative co-occurrences are above a certain threshold, i.e. the pair $(w_i, w_j)$ was added to the vocabulary as a new word if

$$\frac{\text{count}(w_i, w_i) - \delta}{\text{count}(w_i)\text{count}(w_i)} > threshold$$

where $\delta$ is a smoothing term that prevents from forming phrases with rare words. A single pass of this phrase detection method allows to form phrases with pairs of words ("bigrams"); with a second pass, we could also detect relevant phrases with three words ("trigrams").

In practice, we have used a discount factor $\delta=50$, and a threshold of 100 for the first pass (bigrams detection) and 50 for the second pass (trigrams detection).

## 3   Typographical error correction

We have used a simple text correction method to handle typographical errors and misspellings. For this we have relied on the Levenshtein edit distance[4] between pairs of words, that is the minimum number of letters insertions, deletions or substitutions needed to transform one of the words into the other. The edit distance is smaller for similar words (e.g. "cure" and "care" have an edit distance of 1, as a single substitution is needed to transform one word into the other) and large for dissimilar words (e.g. "disease" and "treatment" have an edit distance of seven). A word is considered to be a typo if its occurrences are scarce and if it has a small edit distance with a frequent word. For all pairs of words { $w_r$, $w_f$ } where $w_r$ is a rare word (less than ten occurrences in our corpus) and $w_f$ is a frequent word (more than 100 occurrences), we have replaced $w_r$ with $w_f$ whenever the ratio of their edit distance and the length of the rare word was lower than a given threshold, i.e. when:

$$\frac{\mathrm{d}(w_r, w_f)}{\mathrm{len}(w_r)} > threshold$$

Note that this correction was not performed on small words with less than four letters, as it may lead to unwanted substitutions of many anagrams.